\documentclass[letterpaper, 10 pt, conference]{ieeeconf}  

\IEEEoverridecommandlockouts                              

\usepackage{graphics} 
\usepackage{epsfig} 
\usepackage{times} 
\usepackage{amsmath} 
\usepackage{amssymb}  
\usepackage{amsfonts}
\usepackage{multirow}
\usepackage{breqn}
\usepackage{xcolor}
\usepackage{algorithm}
\usepackage{algorithmic}
\usepackage{mathrsfs}
\usepackage{booktabs} 

\title{\LARGE \bf
Perceive What Matters: Relevance-Driven Scheduling for Multimodal Streaming Perception
}

\author{Dingcheng Huang, Xiaotong Zhang, Kamal Youcef-Toumi
\thanks{All authors are with the Mechatronics Research Laboratory, Massachusetts Institute of Technology, Cambridge, MA, 02139, USA.
{\tt\small \{dean1231,kevxt,youcef\}@mit.edu }}
\thanks{This research was made possible by the support and partnership
of King Abudlaziz City for Science and Technology
(KACST) through the Center for Complex Engineering
Systems at Massachusetts Institute of Technology (MIT) and
KACST.}
}

\begin{document}

\maketitle
\thispagestyle{empty}
\pagestyle{empty}

\begin{abstract}
In modern human-robot collaboration (HRC) applications, multiple perception modules jointly extract visual, auditory, and contextual cues to achieve comprehensive scene understanding, enabling the robot to provide appropriate assistance to human agents intelligently. While executing multiple perception modules on a frame-by-frame basis enhances perception quality in offline settings, it inevitably accumulates latency, leading to a substantial decline in system performance in streaming perception scenarios. Recent work in scene understanding, termed Relevance, has established a solid foundation for developing efficient methodologies in HRC. However, modern perception pipelines still face challenges related to information redundancy and suboptimal allocation of computational resources. Drawing inspiration from the Relevance concept and the information sparsity in HRC events, we propose a novel lightweight perception scheduling framework that efficiently leverages output from previous frames to estimate and schedule necessary perception modules in real-time based on scene context. The experimental results demonstrate that the proposed perception scheduling framework effectively reduces computational latency by up to 27.52\% compared to conventional parallel perception pipelines, while also achieving a 72.73\% improvement in MMPose activation recall. Additionally, the framework demonstrates high keyframe accuracy, achieving rates of up to 98\%. The results validate the framework's capability to enhance real-time perception efficiency without significantly compromising accuracy. The framework shows potential as a scalable and systematic solution for multimodal streaming perception systems in HRC.
\end{abstract}


\section{Introduction}

Perception is a critical component enabling modern intelligent robots to achieve accurate scene understanding. In human-robot collaboration (HRC) scenarios, robotic agents must precisely perceive both the dynamic environment and human partners to provide appropriate assistance. HRC necessitates continuous execution and coordination of multiple perception modules and foundational models, enabling robots to interpret and respond to complex, dynamic environments intelligently\cite{hernandez2025bayesian,kothari2025enhanced}.

Although substantial research has investigated efficient perception under latency and computational constraints, several open challenges persist. Modern perception pipelines typically execute modules frame by frame to optimize perception quality. While activating all available modules every frame ensures accurate offline perception, the computational costs introduce significant latency in streaming settings. In typical parallel pipelines, modules such as object detection and pose estimation execute concurrently, each producing outputs at its own rate. Heavy modules often run at lower frequency, displaying their most recent outputs until new ones become available \cite{mediapipe}. However, modules are still invoked periodically based on readiness, regardless of whether their outputs contribute new or relevant information. For heavy modules, readiness-driven activation may lead to suboptimal timing, potentially missing critical scene information due to a lack of information-aware scheduling.

Another research line leverages keyframe-based approaches, widely applied in video summarization, and key event detection \cite{robot,reviewkey}. Keyframes identified as important serve as triggers for activating perception modules, concentrating computational resources on critical moments while skipping redundant frames. However, keyframe-based approaches present limitations for real-time streaming perception. First, keyframe selection often relies on access to the full video sequence, making it unsuitable for real-time systems where future frames are unavailable. Second, common keyframe criteria---scene change, motion magnitude, visual saliency---are task-agnostic and often misaligned with the perception system's informational needs, selecting irrelevant frames and causing suboptimal resource use without meaningful gains in situational awareness or decision quality.

Research in efficient perception also aims to reduce cost by directly optimizing perception pipelines, primarily minimizing inference cost at the frame level through adjustment of configuration parameters. Common strategies involve fine-tuning settings, reusing intermediate features, or selectively processing input based on perceived importance. However, these efforts predominantly address cost reduction within individual frames without considering perception module scheduling to minimize overall system load. The decision of whether a module should be activated remains largely unexamined. For instance, when processing less informative frames, reducing model size or input resolution may lower inference time, yet bypassing non-essential frames entirely would more effectively eliminate unnecessary computations. Existing methods thus overlook the higher-level optimization of determining activation necessity, which remains crucial for comprehensive efficiency in multi-modal perception systems.

Inspired by the relevance modeling framework, which emulates the human reticular activating system (RAS) to prioritize perceptual inputs based on task context and human intent \cite{zhang2025relevancehumanrobotcollaboration,11127911,zhang2025thesis}, and motivated by the information sparsity in HRC, we introduce a perception scheduling framework that selectively activates perception modules at each frame. The decision process leverages outputs from previous frames and scene dynamics to assess each module's necessity in real time. When a module is not activated, a lightweight estimation, such as state prediction using motion models or filtering, approximates its output, maintaining perception stream continuity without full computational cost. A reward function governs the scheduling policy by quantifying the trade-off between expected information gain and computational efficiency, enabling resource-aware and context-sensitive perception under real-time constraints.

The proposed framework treats all schedulable perception modules as elements within a shared resource pool referred to as the Perception Toolkit. Each module is associated with a reward function quantifying its utility for the current frame. Although exact reward formulations may vary across modules, all follow the same principle: information gain is estimated using a module-specific function, followed by a penalty reflecting computational cost. Once individual rewards are computed, the scheduler evaluates all possible activation combinations and selects the subset maximizing cumulative reward. To demonstrate effectiveness, the current work focuses on two representative modules: object detection and full-body human pose estimation. However, the scheduling principle's generality allows extension to other perception modules and foundation models, such as vision-language models (VLMs), provided appropriate reward formulations are defined. Most importantly, the framework shows potential to mitigate concurrent contention among modules, accommodating larger module sets within fixed computational budgets.

In summary, this paper presents four key contributions toward efficient and adaptive perception in HRC: (1) We introduce the novel concept of perception scheduling within multi-modal perception systems, addressing the fundamental challenge of optimizing module activation to minimize overall computational load. (2) We propose a novel methodology for estimating module activation utility based on information theory, offering a versatile principle applicable to a wide range of perception modules. (3) We develop a novel evaluation metric for real-time perception scenarios, accounting for latency effects and providing a practical performance benchmark. (4) We validate the proposed framework through comprehensive experiments across various streaming video domains, demonstrating improvements in computational efficiency and inference latency while maintaining perceptual quality for collaborative tasks.

\section{Related Works}
 We identified three main fields of work that shared the same goal of optimizing perception systems: keyframe-based methods, computation-aware efficient perception, adaptive sampling and sensor scheduling. To the best of our knowledge, our work is the first to introduce a perception scheduling framework for HRC perception tasks.

\subsection{Keyframe Detection}
Keyframe-based methods are widely adopted in video summarization, segmentation, and activity recognition to reduce redundancy while preserving semantic content. Early techniques rely on motion cues~\cite{1564125,motion2020}, statistical histogram analysis~\cite{CV201536}, or feature-space clustering~\cite{723655} to select visually distinct or representative frames. Other methods employ optimization-based criteria such as sparse reconstruction~\cite{MEI2015522} to ensure coverage with minimal redundancy. More recent approaches incorporate deep learning, including the unsupervised learning-based approach to detect video highlights~\cite{unsupervised2015}, attention-based keyframe scoring for activity recognition~\cite{9344896}, and a self-supervised learning approach that leverages motion and visual features of videos to detect keyframes~\cite{s20236941}.

Despite their effectiveness in compressing visual information, these methods share several key limitations. Most require access to the entire video or a large temporal context, limiting applicability in streaming or interactive scenarios. Frame selection is driven primarily by visual saliency, redundancy, or event likelihood, without regard for the system's moment-to-moment information demand. Crucially, these approaches do not consider per-frame utility, computational cost, or module-level decision-making. By contrast, our framework enables selective and context-aware activation of perception modules in real time, offering finer control and improved efficiency for task-driven robotic systems.

\subsection{Adaptive and Computation-Aware Efficient Perception}
A number of works have addressed perception efficiency by adapting system parameters or processing strategies based on scene context or computational constraints. Several methods focus on frame-level configuration tuning, such as input resolution, tracking frequency, and network scale, to minimize latency while maintaining acceptable performance~\cite{stream2022, zhang2024does}. Other approaches focus on optimizing perception modes by evaluating their impact on robotic stability control performance ~\cite{stability2023}. Recent approaches leverage deep learning-based frameworks, including prioritizing attention mechanisms for scheduling subframe-level data processing for object detector data workflow optimization ~\cite{9804626}. At a higher level, vision-language frameworks have used language models to determine which perception tools to invoke in response to user-defined goals~\cite{vit2024}.

While effective in reducing computational load, most approaches remain limited to single-module scheduling or task-specific pipelines. Decision mechanisms are typically designed around pre-tuned criteria, fixed switching strategies, or offline optimization, with limited flexibility across modalities. General-purpose frameworks for managing multiple perception modules in a unified, context-sensitive manner are largely absent. The lack of per-frame, module-level utility estimation further constrains opportunities for adaptive and selective perception in dynamic, real-time scenarios.

\subsection{Adaptive Sampling and Sensor Scheduling}
Adaptive sampling and sensor scheduling methods selectively 
acquire observations using information-theoretic measures such 
as entropy reduction and mutual information to minimize state 
estimation uncertainty, often under budget or communication 
constraints~\cite{vitus2012efficient, hollinger2014sampling}.

However, fundamental limitations prevent such methods from 
being directly applied to multi-modal perception scheduling in 
HRC. Adaptive sampling and sensor scheduling assume 
homogeneous information sources contributing the same type of 
measurement to a shared state estimate---an assumption that 
breaks down when perception modules produce qualitatively 
distinct outputs such as bounding boxes versus body keypoints. 
Moreover, sampling decisions are governed entirely by 
observation-level uncertainty, with no mechanism to assess 
relevance to the current task objective. In HRC, where human 
intent evolves dynamically, purely uncertainty-driven 
scheduling fails to distinguish between informative and 
irrelevant perception updates.

\section{Perception Scheduling Framework}
In this section, an overview of the perception scheduling framework is introduced, as shown in Fig. 1.
\subsection{Perception Region Segmentation}
The scheduling process begins by partitioning the scene in a structured manner, with each partition processed independently. In this paper, we focus on three distinct categories: background, object, and human. The segmentation leverages the previous frame's raw RGB image $I_{k-1}$, relevance predictions $R_{k-1}$, and prior perception outputs.

For each segmented object and human region, two attributes are assigned. The \textit{motion status} indicates the presence of motion. The \textit{relevance attribute} is updated using prior relevance determination result and reflects the contribution of the region to the current objective.
\subsubsection{Motion Status Estimation}

The motion status for each object or human image patch is determined via frame differencing-based change detection~\cite{Fisher1999ChangeDI}, chosen for its reduced computational complexity when applied at the per-frame level. Let $\mathbf{\Delta P^i}\in \mathbb{R}^{3 \times W \times H}$ denote the absolute difference of patch region \( i \) between consecutive RGB frames. The RGB differences are first converted to grayscale to better capture perceptual changes. Hence, the grayscale difference for the patch region $i$, $\boldsymbol{\Delta L^i}$, is computed as:

\begin{equation}
\boldsymbol{\Delta L^i} = \boldsymbol{Y} \cdot \boldsymbol{\Delta P^i}
\end{equation}
where $\boldsymbol{Y}$ is the standard Luminance coefficient vector.

The change ratio for patch $i$, $\text{CR}^i$, is then computed as the proportion of pixels within the patch exceeding the intensity threshold, $\zeta$:
\begin{equation}
\text{CR}^i = \frac{\sum \left( \boldsymbol{\Delta L^i} > \zeta \right)}{w^i \cdot h^i}
\end{equation}
where $w^i$ and $h^i$ represents the height and width of patch $i$.

The motion status \( s^i \), representing whether patch \( i \) is moving or stationary, is assigned based on the computed change ratio:
\begin{equation}
s^i =
\begin{cases}
\text{moving}, & \text{if } \text{CR}^i > \epsilon \\
\text{stationary}, & \text{otherwise}
\end{cases}
\end{equation}
where $\epsilon$ is the patch change threshold.

\subsubsection{Relevance State Update}

Each segmented object and human patch is assigned a relevance label using predictions from the relevance framework. Let \( r^i_k \) denote the relevance state of patch i at frame k, representing the conditional probability that the patch is relevant to the human objective. The relevance state is retained as a conditioning variable for downstream reward estimation.

\subsection{Perception Reward Estimation}
For each perception module $m^j \in \mathcal{M}$ in the perception toolkit, a reward $\rho_k^j$ is computed at time $t$ to quantify the utility of activating the module. The estimation relies on the segmentation result from the previous stage, which includes motion status and relevance attributes for each region in the frame. The reward is defined as the trade-off between the expected information gain and the corresponding computational cost penalty:
\begin{equation}
    \rho_k^j = \Phi_R(S_k, m^j) - C^j
\end{equation}
where \( \Phi_R(S_k, m^j) \) denotes the estimated information gain when module \( m^j \) is activated under the current scene state \( S_k \), and $C^j$ is the penalty applied to the latency or computational cost of running $m^j$. This formulation enables the selection of perception modules that provide high contextual benefit relative to their computational cost, promoting efficient resource utilization in real-time settings.
Note that the function \( \Phi_R \) is defined in an independent and module-specific manner to reflect the type of information each module contributes. The methodology section provides a detailed description of the reward estimation process for object detection and human full-body pose estimation.
\subsection{Perception Module Selector}

Given the estimated reward for all modules in the perception toolkit, the optimal scheduling decision at each frame is made by maximizing the accumulated reward over the set of possible module activations. We define the action space $\mathcal{A}$ as the set of all binary activation vectors, where $a^j = 1$ indicates that module $m^j$ is activated and $a^j = 0$ otherwise. The optimal policy $\pi^*[k]$ at frame $k$ is formulated as:

\begin{equation}
\pi^*[k] = \arg\max_{a \in \mathcal{A}} \left( \sum_{j=1}^{n} a^j \cdot \rho_k^j \right) \text{s.t. } a^j \geq G_1^j[k], \; \forall j
\label{eq:selector}
\end{equation}

Where $G_1^j[k] \in \{0, 1\}$ is a predetermined activation indicator for module $j$ 
at time step $k$, representing scheduling decisions that do not depend on 
reward-based assessment. Since the reward for each module is estimated independently, 
the optimization in Eq.~\eqref{eq:selector} decomposes into 
independent per-module decisions. Specifically, module $m^j$ is 
activated if and only if $\rho_k^j > 0$ or $G_1^j[k] = 1$, 
eliminating the need for combinatorial search over the action 
space $\mathcal{A}$.

The execution outputs of the selected optimal set of modules are forwarded to the relevance framework for context-aware relevance determination. The resulting information is then stored and used as part of the previous frame's result, enabling informed reward estimation in subsequent scheduling decisions.

\begin{figure*}[!t]
\begin{center}
\includegraphics[width=0.95\textwidth]{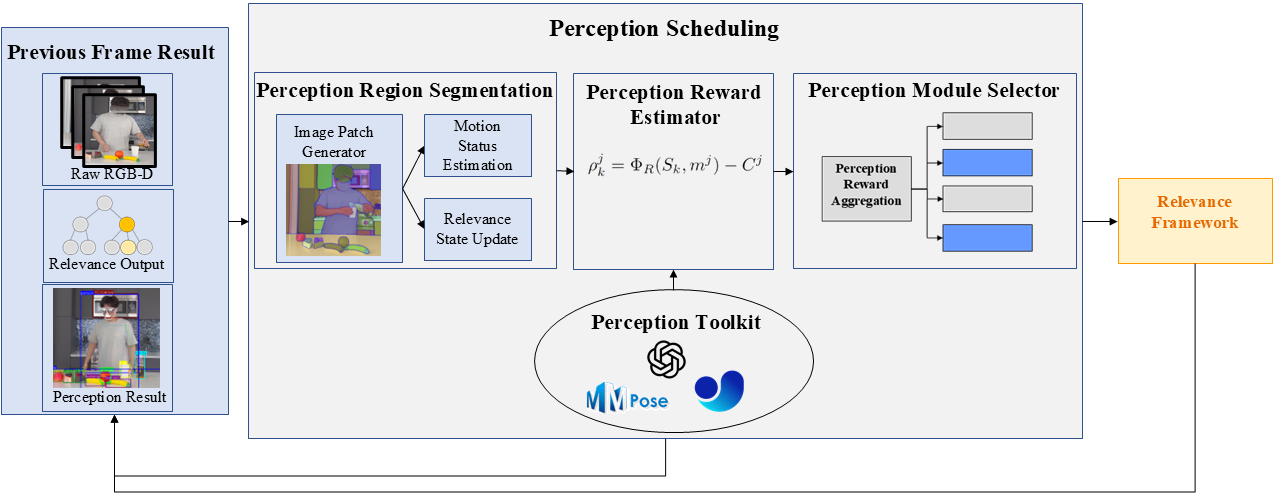}
\end{center}
\caption{Perception scheduling framework for context-aware and efficient perception in human-robot collaboration. The system uses information from the previous frame to segment relevant regions, estimate motion status, and update relevance states. Each module in the perception toolkit is evaluated based on a reward that balances expected information gain and computational cost. The module selector computes the optimal activation set at each frame, and the selected outputs are used to update the relevance framework.
}
\label{fig: framework}
\end{figure*}

\section{Reward Modeling}
In this section, we present a detailed reward modeling process for a scheduling problem that incorporates object detection and human full-body pose estimation.

\subsection{Reward Model for Object Detection}
\label{subsec:reward}

The reward modeling for object detection incorporates two primary sources of information gain: (1) identifying whether scene elements have entered or exited the current view, and (2) updating the state estimates of already tracked elements. A computational penalty is also included to account for the cost of activating the module.

To detect changes in scene composition, change detection is applied to masked background regions. The previously introduced frame differencing method is used to capture localized pixel-level variation, while color histogram analysis is employed to detect global shifts in appearance. The histogram distance is calculated per RGB channel using the Chi-Square metric, with the total shift, $D_H$, given by the mean across channels. If both the histogram change, $D_H$, and the previously defined change ratio exceed their respective thresholds, the module activation indicator, $G_1^{yolo}[k]=1$, is assigned to reflect the discovery or removal of scene elements at the current frame.

The reward from updating existing element states is computed based on the entropy reduction in the bounding box estimates. In order to estimate the bounding box states before turning on the object detection module, we adopted the Kalman filter implementation in BoT-SORT \cite{botsort}. The bounding box state in BoT-SORT is represented as an 8-dimensional state vector, which encapsulates both the spatial properties of the bounding box and the dynamic motion state.

We perform Kalman filter state prediction under the constant velocity assumption. Note that the uncertainty estimation is performed based on the predicted state uncertainty. In cases where the object detection module is not activated, we propagate the prediction without the Kalman filter update. If object detection is scheduled to be activated for the current frame, the update is performed to reflect the effect of the measurement on the bounding box state. Therefore, the resulting information gain for the object detection $G_2^{\text{yolo}}[k]]$ at the current frame is:
\begin{align}
G_2^{\text{yolo}}[k] =\ &  \sum_{p=1}^{n} \frac{1}{2} r^p_k \log\left(\frac{\det(H \mathcal{\bar{P}}^p_k H^\top)}{\mathcal{\det(R)}}\right) 
\end{align}

\noindent where $\mathcal{\bar{P}}_k$ is the prior covariance in state space at the current frame, $H$ is prior covariance in the measurement space, $\mathcal{R}$ is the measurement uncertainty of the object detection module, $n$ is the number of scene elements, and $r^p_k$ is the relevance of the scene element $p$.

The reward for object detection at frame $k$ is given by:
\begin{equation}
\rho_k^{\text{yolo}} = G_2^{\text{yolo}}[k] - \lambda C^{\text{yolo}}
\end{equation}

\noindent where $C^{\text{YOLO}}$ denotes the standard inference time of running the object detection module in milliseconds, and $\lambda$ is the Lagrange multiplier with the unit of $bits/ms$ that represents the information value of the inference time.

\subsection{Reward Model for Human Full-body Pose Estimation}
The reward modeling for human full-body pose estimation in the perception scheduling framework captures the trade-off between information gain and computational cost following the general principle discussed above.

The information gain from running MMPose consists of two main components. The first component captures changes in scene composition arising from humans entering or exiting the workspace. The second component is the information gain obtained from executing the MMPose module itself. This component quantifies the reduction in uncertainty when transitioning from knowledge about the bounding box location of the human agent to the keypoint locations of the human agent. In other words, it measures how much more precise the scene understanding becomes when keypoints are detected rather than just the human bounding box.

For the first component, we apply the same change detection method described in Section~\ref{subsec:reward}, combined with human tracking results, to detect human-agent scene composition changes.

For the second component, an uncertainty model is developed to quantify the uncertainty of human pose knowledge based on the bounding box location. To model the uncertainty of a uniformly distributed keypoint within a bounding box of width $w$ and height $h$, we derive the entropy of a continuous uniform distribution over the region. 

For a uniform distribution over a rectangular region of area $A = w \cdot h$, the probability density function is computed as:

\begin{equation}
p(x, y) = \frac{1}{w h}, \quad \text{for } x \in [0, w],\ y \in [0, h]
\end{equation}

The differential entropy $\mathcal{H}$ is given by:

\begin{align}
\label{eq:my-eq}
\mathcal{H} &= - \int_0^w \int_0^h \frac{1}{w h} \ln\left( \frac{1}{w h} \right) dy\, dx = \ln(w \cdot h)
\end{align}

Thus, Eq.~\eqref{eq:my-eq} represents the entropy of a uniformly distributed keypoint within the bounding box.

The aggregated pre-execution uncertainty across all keypoints is therefore defined as:

\begin{equation}
\mathcal{H}^{\text{pre}}_{k} = D\sum_{s=1}^{N} r^s\left( \ln[(w_{k}^{s}+{\sigma}^s_w)( h_{k}^s+{\sigma}^s_h)] \right)
\end{equation}

\noindent where \( w_k^s \) and \( h_k^s \) denote the width and height of the human bounding box s at frame \( k \), \( \sigma_{w,k}^s \) and \( \sigma_{h,k}^s \) represent the corresponding Kalman prediction uncertainty in prior covariance matrix, $D$ is the number of keypoints, and $N$ is the number of human. In full-body human pose estimation, each human contains 133 keypoints.

To model the estimated uncertainty of the human pose estimation output, we first introduce the confidence score $s_d^k \in (0,1]$, which represents the model's confidence in the predicted position of keypoint $d$ at frame $k$. The confidence score provides a direct indication of the reliability of the estimated keypoint location. 

Next, we define the uncertainty associated with each keypoint. Let $D$ denote the total number of keypoints in the pose skeleton. For each keypoint $d$, we model the predictive uncertainty as a two-dimensional random variable over its $(x, y)$ position. Assuming the independence between the coordinates $x$ and $y$ and the symmetric uncertainty for the coordinates $x$ and $y$, the uncertainty in frame $k$ is captured using a diagonal covariance matrix $\Sigma^d_k$:

\begin{equation}
\Sigma^d_k =
\begin{bmatrix}
\mathrm{Var}^{d}_k & 0 \\
0 & \mathrm{Var}^{d}_k
\end{bmatrix}
\end{equation}

Given the confidence scores $s_d^{k_{\text{last}}}$ and $s_d^{k_{\text{prev}}}$ 
obtained from previous model executions for keypoint $d$, where $k_{\text{last}}$ 
and $k_{\text{prev}}$ denote the most recent and second most recent frames at 
which the model was executed, we linearly extrapolate the estimated confidence 
score for the current frame $k$  as:

\begin{equation}
\hat{s}^d_k = \text{clamp}\left(s^d_{k_{\text{last}}} + \frac{s^d_{k_{\text{last}}} - s^d_{k_{\text{prev}}}}{k_{\text{last}} - k_{\text{prev}}} \cdot (k - k_{\text{last}}), \; \epsilon, \; 1\right)
\end{equation}

where the result is clamped to the interval $[\epsilon, 1]$ to ensure 
the extrapolated score remains a valid confidence value, with $\epsilon$ 
being a small positive constant to prevent numerical instability.

The standard deviation $\sigma^d_k$ of keypoint $d$ at the current frame is then estimated from the confidence score using a negative log mapping function:

\begin{equation}
\sigma^d_k = -\sigma_{base}^d\log(\hat{s}^{d}_k)
\end{equation}

where $\sigma_{base}^d$ is the base uncertainty, derived from the per-key point standard deviation from COCO dataset annotations and the object scale

Therefore, the entropy of the 2D Gaussian distribution for keypoint $d$ at the current frame is given by:

\begin{equation}
\mathcal{H}^d_k = \frac{1}{2} \log\left( (2\pi e)^2 \cdot \det(\Sigma^d_k) \right) = \log(2\pi e) + 2\log(\sigma^{d}_k)
\end{equation}

Here, $\mathcal{H}^d_k$ represents the estimated entropy of keypoint $d$ at the current frame for a single human agent, quantifying the uncertainty in the spatial estimation.

To generalize the formulation for multi-agents with the relevance weighting, the total entropy across all keypoints at the current frame is then given by:
\begin{align}
\mathcal{H}^{\text{pose}}_k &= \sum_{s=1}^{N} r^s_k \sum_{d=1}^{D} \mathcal{H}^{d,s}_k \\
&= \sum_{s=1}^{N} r^s_k [D \log(2\pi e) +  \sum_{d=1}^{D} 2\log\left( \sigma^{d,s}_k \right)]
\end{align}
where $H^{d,s}_k$ represents the entropy of keypoint $d$ of human $s$ at the current frame, $\sigma^{d,s}_k$ represents its corresponding standard deviation, and and $r^s_k$ is the relevance of the human agent s.

In summary, the total reward for full-body pose estimation at the current frame is:
\begin{equation}
\rho_k^{\text{pose}} = G_2^{\text{pose}}[k] - \lambda C^{\text{pose}}
\end{equation}
where  $C^{\text{pose}}$ is the standard inference time of MMpose, and $G_2^{\text{pose}}[k]$ is defined as:
\begin{equation}
G_2^{\text{pose}}[k]=\mathcal{H}_k^{\text{pre}}-\mathcal{H}_k^{\text{pose}}
\end{equation}

\section{Experimental Setup}
To demonstrate the effectiveness of the perception scheduling framework, we propose a novel evaluation metric and conduct experiments on three distinct video domains.
\subsection{Hardware Setup}
The evaluation videos are recorded at 30 FPS to emulate the standard frequency of a streaming perception system pipeline. The inference of the perception module is performed on an RTX 4090 GPU, while perception scheduling is executed on an Intel i9 CPU to ensure efficient real-time processing.

\subsection{Video Domains}
Since existing datasets do not contain ground-truth information about relevance states in the scene, we curated three videos representing different domains of human daily activity to perform comprehensive evaluations on the perception scheduling framework across varied scenarios.

\textbf{Indoor Reading}: In this domain, a human enters the scene and engages in reading activities, thereby capturing static, low-movement interactions typical of indoor environments.

\textbf{Eating}: The Eating domain features a human performing eating-related activities. The domain setup reflects moderate human movement and frequent human-object interaction.

\textbf{Walking}: The Walking domain represents an outdoor environment where the human walks through the scene with higher motion levels.

\subsection{Perception Module Selector Evaluation}
To rigorously assess the effectiveness of our perception scheduling framework, we introduce two key evaluation metrics: \textbf{latency} and \textbf{Activation Recall}. Latency measures the computational efficiency of the system and is defined as the average processing time per frame (in milliseconds) over a given duration. Specifically, it encompasses both the module scheduling processing time and the module inference time to reflect the level of complexity for a perception task. To calculate latency, we compute the total time taken to schedule and execute the perception modules (YOLO and MMPose) and divide it by the number of keyframes processed.

The latency metric reflects the system's computational efficiency and provides a quantitative representation of the computational cost savings achieved through the perception scheduling framework.

The activation recall is defined as the percentage of frames where a perception module was activated when it was genuinely necessary. The metric specifically evaluates the performance of our scheduling decisions rather than the intrinsic accuracy of the perception modules. 

To establish the ground truth, we conduct an offline evaluation by running both perception modules on every frame without scheduling. The resulting detection outputs are used to identify frames that actually require module activation. Our recall metric is then computed as:

\begin{equation}
\text{Recall} = \frac{\text{Number of Correctly Activated Frames}}{\text{Number of Frames Requiring Activation}}
\end{equation}

We evaluate the performance of the perception scheduling framework against two baseline benchmarks: \textbf{Conventional Parallel Perception Pipeline} and \textbf{Oracle Scheduling}. The \textbf{ conventional parallel perception pipeline} represents a standard perception setup where the perception modules (YOLO and MMPose) run in parallel. In this configuration, the perception modules are activated as soon as the previous inference queues have been processed. The conventional approach does not incorporate any scheduling logic, thereby utilizing the perception modules continuously whenever computational resources permit.

The second benchmark, \textbf{Oracle Scheduling}, directly leverages the ground-truth information obtained from offline processing. The module activation strictly follows the ground truth frames that require the perception module execution. Essentially, Oracle Scheduling is an idealized system that assumes perfect knowledge of the frames that necessitate activation, without any decision-making. The setup serves as upper bounds for scheduling accuracy and efficiency.

\begin{table}[]
\centering
\caption{Perception Scheduling Framework Performance Evaluation. Our framework Effectively Reduces Latency by up to 27.52\% and improves activation recall by up to 72.73\% Under Inherent Perception Delay.}
\begin{tabular}{ccccc}
\hline
Domains                        & Metrics & Parallel     & \begin{tabular}[c]{@{}c@{}} Oracle \end{tabular} & \begin{tabular}[c]{@{}c@{}} Perception \\ Scheduled \end{tabular} \\ \hline
\multirow{3}{*}{Indoor Reading} & Latency (ms)    & 98.81 & 48.95                                                    & \textbf{71.62}\\
                               & Yolo Recall & \textbf{1.00}  & 1.00                                                     & 0.97\\
                               & Pose Recall    & 0.16           & 0.24                                            & \textbf{0.20}                                                       \\ \hline
\multirow{3}{*}{Eating} & Latency (ms)  &94.99  & 70.00                                                   &\textbf{86.44}\\
                               & Yolo Recall & \textbf{1.00}  & 1.00                                                     & 0.98\\
                               & Pose Recall    & 0.16           & 0.24                                            & \textbf{0.20}                                                      \\ \hline
\multirow{3}{*}{Walking}   & Latency (ms)   & 93.63 & 81.51                                                       & \textbf{75.15}                                                       \\
                               & Yolo Recall & \textbf{1.00}  & 1.00                                                     & 0.93                                                        \\
                               & Pose Recall    & 0.22          &   0.52                                          &     \textbf{0.38}                                                   \\ \hline
\end{tabular}
\label{tab:1}
\end{table}

The experimental results are shown in Table \ref{tab:1}. Our proposed perception scheduling framework demonstrates a significant reduction in latency compared to the parallel benchmark, achieving up to a 27.52\% reduction. The improvement highlights the effectiveness of the perception scheduling mechanism in reducing computational costs. Latency reduction is most pronounced in relatively static, low-complexity settings (e.g., Indoor Reading). From an information-theoretic perspective, predicted states align with observations, yielding limited information gain. Similarly, entropy estimation for human tracking in such scenarios provides little incentive to activate perception modules. Consequently, the computational cost of activation often exceeds the marginal information gain, leading to more conservative scheduling of perception modules.

In contrast, more dynamic scenarios, such as the Eating and Walking domain, exhibit a less substantial reduction in latency. The increase in the number of frames requiring module activation, due to frequent motion and scene changes, directly impacts the scheduling efficiency. In such cases, the state estimates become less reliable, and the entropy associated with human tracking rises, increasing the information gain when perception modules are activated. As a result, the perception scheduling framework identifies more frames as requiring activation, which reduces the latency advantage. 

Another key observation from the results is the improvement in MMPose activation recall in our proposed framework compared to the parallel benchmark. In particular, the proposed framework achieves the best MMPose recall (38\%) and improves the most in the Walking domain (72.73\%) as dynamic scenes increase the occurrence of activation-critical frames that conventional pipelines fail to capture under perception delay. Despite the observed improvement, the MMPose recall remains relatively low across all domains. The outcome is attributed to the inherent inference latency of the full-body pose estimation module. Even when the scheduling mechanism accurately activates the module on necessary frames, the processing delay results in intermediate important frames being missed when they occur during the inference. For instance, when MMPose is activated at frame 10 and completes processing by frame 14, any important frames appearing in between are not captured. Nevertheless, despite the relatively low MMPose accuracy, the perception scheduling framework still manages to significantly reduce the number of important frames being missed. 

The final important observation from the results is the slight reduction in YOLO recall across all video domains when using the perception scheduling framework. The minor decrease stems from the inherent accuracy limitations of the perception modules and the uncertainty involved in estimating information gain. However, the overall performance improvement and latency reduction achieved by the perception scheduling framework outweigh the drop in YOLO recall. The YOLO scheduling still maintains comparable performance to the benchmark methods, demonstrating that the efficiency gains do not significantly compromise the perception quality.

\begin{table}[]
\centering
\caption{Perception Scheduling Framework Keyframe Identification Performance Evaluation. Our Methodology Can Accurately Predict The Important Frames To Be Scheduled }
\begin{tabular}{ccc}
\hline
Domains                       & \begin{tabular}[c]{@{}c@{}} Yolo Keyframe \\  Accuracy\end{tabular} & \begin{tabular}[c]{@{}c@{}} MMpose Keyframe \\  Accuracy\end{tabular}\\ \hline
\multirow{1}{*}{Indoor Reading} & 0.97 & 0.89\\ \hline
\multirow{1}{*}{Eating}  & 0.98 & 0.97\\ \hline                                                 
\multirow{1}{*}{Walking}  & 0.93 & 0.92 \\ \hline
\end{tabular}
\label{tab:2}
\end{table}

\subsection{Reward Model Evaluation}
To evaluate the performance of the reward model, we define a new metric termed \textbf{Keyframe Accuracy}. The \textbf{Keyframe Accuracy} is defined as the percentage of important frames that are correctly identified for activation. This metric specifically evaluates the perception scheduling framework’s ability to detect which frames require processing, independent of whether the processing itself was executed. In contrast, the scheduling accuracy discussed earlier measures the percentage of important frames that were actually processed. By focusing on keyframe accuracy, the analysis isolates the effect of the inherent latency of perception modules, providing a more precise assessment of the framework's capability to identify critical frames.

As shown in Table \ref{tab:2}, our framework demonstrates high Keyframe accuracy across all domains. The result indicates that our framework can effectively determine when to activate perception modules in streaming perception settings, independent of the latency introduced by the modules.

\section{Demonstration Results}
\begin{figure}
\begin{center}
\includegraphics[width=0.43\textwidth]{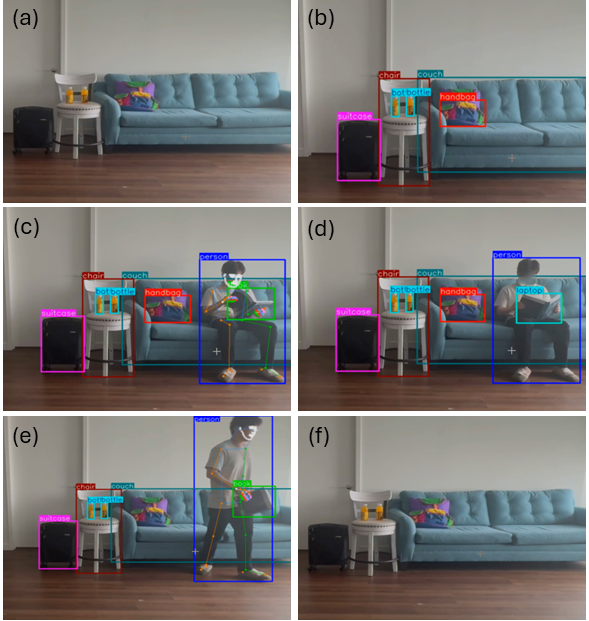}
\end{center}
\caption{Demonstration of perception scheduling. The Perception scheduling framework effectively adapts module activation to scene context
}
\label{fig: demo}
\end{figure}

Fig. \ref{fig: demo} illustrates the adaptive behavior of the perception scheduling framework across different scene contexts demonstrated in the video. In Fig. 2(a–b), under a static scene without a human, YOLO is activated at a frequency governed by estimated information gain rather than every frame, reducing redundant computation.

When a human enters the scene (Fig. 2(c)), YOLO activates and MMPose is triggered to estimate the human pose. In Fig. 2(d), the human remains seated and relatively motionless while reading, prompting the framework to reduce MMPose activation frequency to save computational resources without degrading perception quality.

In Fig. 2(e), the human begins leaving the scene, causing a contextual change that activates both YOLO and MMPose continuously for accurate tracking. Finally, Fig. 2(f) depicts the post-exit scene. MMPose remains inactive, and YOLO is scheduled in a manner similar to the Fig. 2(a–b) scenario.

\section{Conclusion}
In this paper, we presented a novel perception scheduling 
framework designed to enhance the efficiency of multi-modal 
perception in human-robot collaboration (HRC) settings. The 
proposed framework leverages the concept of relevance, 
dynamically activating perception modules based on estimated 
information reward and computational cost.

Our experimental results demonstrate the effectiveness of the 
proposed framework. Compared to conventional parallel 
perception pipelines, the framework achieved up to 27.52\% 
reduction in latency, a 72.73\% improvement in MMPose 
activation recall, and keyframe accuracy rates up to 98\%, 
validating its ability to optimize computational resources 
while maintaining perceptual quality.

The modular nature of the framework allows for incorporation 
of additional perception modules beyond object detection and 
pose estimation. Extending the scheduling principle to heavier 
modules such as vision-language models (VLMs) 
requires adapting the reward formulation to capture their 
unique information characteristics---for instance, modeling 
the semantic information gain of a VLM query relative to 
existing scene context, while accounting for substantially 
higher and more variable inference costs. The core scheduling 
mechanism remains applicable, as the reward-cost trade-off 
generalizes regardless of module complexity.

Future work will focus on developing adaptive methods across diverse scenarios and extending 
the framework to incorporate resource coupling constraints 
when scheduling larger sets of perception modules under 
shared computational budgets~\cite{hu2025eye}. 

\addtolength{\textheight}{0cm}   




\bibliographystyle{IEEEtran}
\bibliography{IEEEabrv,references}

\end{document}